# Mining and Discovering Biographical Information in *Difangzhi* with a Language-Model-based Approach

Peter K. Bol, Harvard University (USA), peter_bol@harvard.edu

Chao-Lin Liu, National Chengchi University (Taiwan), chaolin@nccu.edu.tw

Hongsu Wang, Harvard University (USA), hongsuwang@fas.harvard.edu

## Background

We present results of expanding the contents of the China Biographical Database[1] by text mining historical local gazetteers, *difangzhi*地方志 [2]. The goal of the database is to see how people are connected together, through kinship, social connections, and the places and offices in which they served. The gazetteers are the single most important collection of names and offices covering the Song through Qing periods. Although we begin with local officials we shall eventually include lists of local examination candidates, people from the locality who served in government, and notable local figures with biographies. The more data we collect the more connections emerge. The value of doing systematic text mining work is that we can identify relevant connections that are either directly informative or can become useful without deep historical research. Academia Sinica is developing a name database for officials in the central governments of the Ming and Qing dynasties[3].

## Problem Definition and Main Findings

Figure 1 shows a scanned page from a *difangzhi*[4] and Figure 2 shows the text of the shown image. As Figure 1 shows traditional Chinese texts do not have spaces between words or employ punctuation. This feature makes the processing of literary Chinese texts much more difficult than handling alphabetical languages and modern Chinese. The circles in Figure 2 serve as a general delimiter, representing the end of a line, the end of a page, a space, or a transition in text formatting, e.g., placing two lines of text in a single column in column 2 of Figure 2.

     We would like to algorithmically extract the information about local officials, such as Li Chang 李常 in Figure 2. We are interested in the alternative names, such as the style name (*zi* 字) and pen name (*hao* 號), birth place, entry method, serving office, service time, etc. In this abstract, we focus on how we identify a person's name, style name, and the dynasty. For

---

[1] <http://isites.harvard.edu/icb/icb.do?keyword=k16229>The China Biographical Database (CBDB) is a collaborative project of Harvard University, Peking University, and Academia Sinica. CBDB is an online relational database with biographical information about approximately 328,000 individuals as of October, 2013, primarily from the 7th through 19th centuries. The data is meant to be useful for statistical, social network, and spatial analysis as well as serving as a kind of biographical reference.

[2] *Difangzhi*: <http://www.chinaknowledge.de/Literature/Terms/difangzhi.html>

[3] http://archive.ihp.sinica.edu.tw/ttsweb/html_name/search.php

[4] When written in the vertical style, Chinese paragraphs begin from the right side of a page.

example, we wish to extract the record Song 宋 as dynasty, Li Chang 李常 as person name, and Gongze 公擇 as style name after handling the text in Figure 2.

Local gazetteers record various types of information about local areas; we selected those that are related to local government officials. Not counting the circles in the texts, the current study employed 83 text files containing 901,302 Chinese characters.

We extracted 1260 records from the files, and compared them with the biographical data in CBDB. Table 1 gives an analysis of the results, where a circle indicates a match, and a cross a mismatch. Among the 1260 records, 562 match the dynasty, personal name, and style name of some CBDB records, and 544 (43.2%) match only dynasty and name.

**Methods**

Figure 3 shows the main procedure for extracting the records. In addition to the gazetteers, we used files of previously known names, addresses, entries, offices, and reign period titles from CBDB to annotate the texts[5]. In this study, we also consider the dynasties for names, offices, and reign periods.

We need to consider the ambiguities of a word when annotating the texts. For instance, "Li Chang" 李常 was a person name in the Song, Yuan, Ming, and Qing dynasties, and the four-character office title "Guan-cha tui-guan" 觀察推官 was an office in the Tang and Song dynasties. In addition the second and third characters "cha tui" 察推 was also an office in the Song and Yuan dynasties. Hence, as illustrated in Figure 4, we could generate at least 16 possible label sequences for the following string T1 in Figure 2.

T1: 李常字公擇南康建昌人自宣州觀察推官發運使

We sift the label sequences by adopting the principle of favoring longer words[6] and by disambiguating with contextual constraints. In T1, we do not consider "cha tui" 察推 an office for the Song dynasty because the four character sequence is a longer match for the same dynasty. In addition, it is reasonable to require that all labels in a sequence must be ***consistent*** with the same dynasty. Hence, among the 16 sequences, only "李常-觀察推官" for Song and "李常-察推" for Yuan could survive.

Since there are no known tools for parsing literary Chinese, we employ the concept of language models (Manning & Schütze, 1999) to analyze the texts. We computed, collected, and counted the frequencies of ***consistent sequences*** of six labels[7].

---

[5]Here, "names" include either official names or any alternative names. "Addresses" refer to location names. "Entries" (入仕方式), e.g., "進士" and "舉人", include different ranks and ways of becoming a government official via the Civil Service Examinations ("科舉"). "Offices" (官職, government positions) include posts in the government. "Reign period names" (年號), e.g., Kangxi 康熙, are names of time periods under a particular emperor.
[6]This so-called "favoring the longer term" 長詞優先 principle is commonly adopted when segmenting (or tokenizing) Chinese text strings, cf. (Gao et al., 2005)
[7]Technically speaking, we are analyzing a 6-gram language model.

Aiming at extracting personal names and style names for government officials, we focused on the consistent sequences that have at least one <NAME> label. We then identified and preferred subsequences that include more different labels. We show four such ***filter patterns*** below.

P1: <NAME><ADDRESS><REIGN PERIOD><ENTRY>
P2: <NAME><ADDRESS><ENTRY><REIGN PERIOD>
P3: <NAME><NAME><ADDRESS><ADDRESS>
P4: <NAME><ADDRESS><ADDRESS><ADDRESS>

Finally, we selected the consistent sequences which contained the filter patterns, and extracted original text segments that corresponded to the consistent sequences. The string T1 was extracted from the text in Figure 2 because it could be annotated with the sequence <NAME><ADDRESS><ADDRESS><ADDRESS><OFFICE><OFFICE>, which contained P4. The <NAME> label is for "李常". At the annotation stage, our programs did not recognize "Gongze" 公擇 as a Style Name for "Li Chang" 李常 because "Gongze" 公擇 was not included in the CBDB name list. One of the two annotated results is listed below.

**<NAME Song>李常</NAME>字公擇<ADDRESS>南康</ADDRESS><ADDRESS>建昌</ADDRESS>人自<ADDRESS>宣州</ADDRESS><OFFICE Song>觀察推官</OFFICE><OFFICE Song>發運使</OFFICE>**

To extract "Gongze" 公擇 as a Style Name from T1, we parsed the text segment with a low-level grammar pattern for the task. Specifically, a two-character string that appears after the sequence of a <NAME> label and the character "字" (Style Name) and before an <ADDRESS> label was extracted as the Style Name for the <NAME>. With such syntactic rules, we *discovered* that "公擇" is a Style Name for "李常", and obtained two records: (Song,李常,公擇) and (Yuan,李常,公擇).

## Results, Evaluation, and Applications

We compared the extracted records with the combinations of dynasty, name, and Style Name in CBDB, and Table 1 shows the results. The two records that we just obtained would belong to type 2, because "Gongze" 公擇 is not known to CBDB. All extracted records of type 2 provide opportunities of finding Style Names that were new to CBDB. However, they should be confirmed by asking a domain expert to check the original text segments, which is an operation facilitated by our software platform.

Extracted records of type 1 do not provide new information if we are just interested in names and Style Names. Certainly, we are more ambitious than this, and type-1 records are instrumental. They help us find the beginnings of the paragraphs that contain extra information about the owners of the type-1 records. T1 is the beginning of the second paragraph in Figure 1. This paragraph contains extra information about "李常" that we can explore to enhance the contents of CBDB. The third paragraph in Figure 1 and many following paragraphs start with statements that we could identify with the filter patterns.

Records of types 3 through 7 make only about 12.2% of the 1260 extracted records. Similar to type-2 records, these records do not match any records in CBDB perfectly. After inspecting the original text segments, we will be able to tell whether these mismatches are new discoveries or incorrect extractions.

## Discussion

The reported work represents an extension of our work for CBDB that was reported in (Bol et al., 2012). In the previous work, experts manually designed regular expressions for specific text patterns. Now, based on prior information about named entities, we are able to compute and analyze the label sequences for the local gazetteer texts to learn useful filter patterns for automatically extracting desired information. We can apply the reported mechanism to extract birth places, service periods, offices, and other basic information, as we just did for extracting names and Style Names. In addition, by identifying key opening statements for paragraphs that contain biographical data, the reported procedure opens a new door for algorithmically extracting information about personal career and social networks. We are working toward learning the document structures of local gazetteers.

Our work is related to automatic grammar induction in computational linguistics. Hwa (1999) learns grammars with data that were manually annotated with syntactic information, and we automatically annotated data with named entities. Klein and Manning (2005) employed advanced techniques to learn hierarchical grammars for Penn treebank sentences, which may be quite challenging in the case of literary Chinese.

Additional responses to reviewers' comments are available online at the following URI: <http://www.cs.nccu.edu.tw/~chaolin/papers/dh2015blw.online.pdf>.

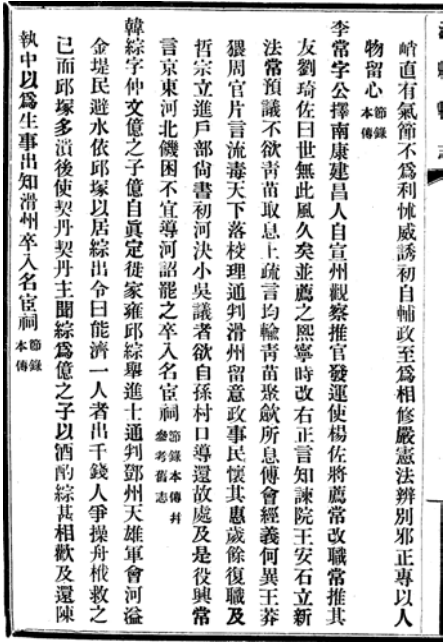

**Figure 1.** A scanned image of a page of the *Difangzhi* books

○○峭　有氣節不爲利怵威誘初自輔政至爲相修嚴憲法辨別邪正專以人○○○物留心○節錄○本傳○李常字公擇南康建昌人自宣州觀察推官發運使楊佐將薦常改職常推其○○○友劉琦佐曰世無此風久矣並薦之熙寧時改右正言知諫院王安石立新○○○法常預議不欲青苗取息上疏言均輸青苗聚歛所息傅會經義何異王莽○○○猥周官片言流毒天下落校理通判滑州留意政事民懷其惠歲餘復職及○○○哲宗立進戶部尚書初汭小吳議者欲自孫村口導還故處及是役興常○○○言京東河北饑困不宜導河詔罷之卒入名宦祠○節錄本傳并○參考舊志○韓綜字仲文億之子億自眞定徙家雍邱綜舉進士通判鄧州天雄軍會河溢○○○金堤民避水依邱塚以居綜出令曰能濟一人者出千錢人爭操舟枚救之○○○已而邱塚多潰後使契丹契丹主聞綜爲億之子以酒酌綜甚相歡及還陳○○○執中以爲生事出知滑州卒入名宦祠○節錄○本傳○

**Figure 2.** A passage for the image in Figure 1

**Table 1.** Analysis of the extracted 1260 records

| Type | Dynasty | Name | Style Name | Quan. | Prop. |
|---|---|---|---|---|---|
| 1 | ○ | ○ | ○ | 562 | 44.6% |
| 2 | ○ | ○ | × | 544 | 43.2% |
| 3 | × | ○ | ○ | 40 | 3.17% |
| 4 | ○ | × | ○ | 31 | 2.46% |
| 5 | × | ○ | × | 29 | 2.30% |
| 6 | × | × | ○ | 20 | 1.59% |

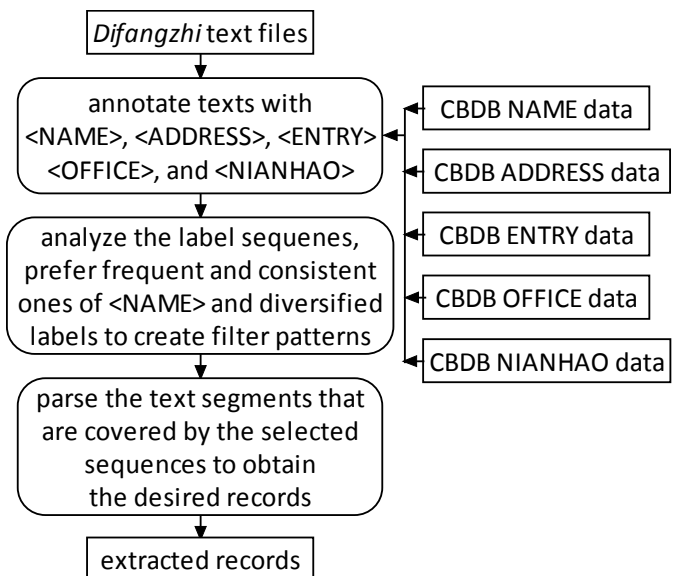

**Figure 3.** Main steps of the extraction procedure

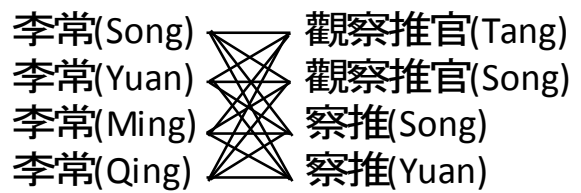

**Figure 4.** Preferring longer words and context-based disambiguation

# Online Discussions

We itemize our responses to some reviewers' questions about the following DH2015 paper.

**Peter K. Bol, Chao-Lin Liu, and Hongsu Wang.** (2015). Mining and discovering biographical information in Difangzhi with a language-model-based approach, *Proceedings of the 2015 International Conference on Digital Humanities*, 000–000. Sydney, New South Wales, Australia, 29 June-3 July 2015.

## Reviewer 1

**1. The most important characteristics about local gazetteers is the locality. The paper did not mention the locations of the 83 gazetteers to be explored, nor the respective time period of their completion.**

Here is the list of the 83 books which we used in this study and their completion periods. The digits in the parentheses indicate the number of volumes of the corresponding sources.

| Source | Period | Source | Times | Source | Times | Source | Times |
|---|---|---|---|---|---|---|---|
| 順德縣志 | 民國 | 豐縣志 | 光緒 | 祁門縣志 | 同治 | 太湖縣志 | 民國 |
| 陽江志 | 民國 | 華亭縣志 | 乾隆 | 慈利縣志 | 民國 | 續修舒城縣志 | 光緒 |
| 臨晉縣志 | 乾隆 | 青浦縣志 | 光緒 | 長沙府志 | 乾隆 | 廣西通志輯要 | 光緒 |
| 漢州志(3) | 嘉慶 | 樂昌縣志 | 民國 | 襄城縣志 | 乾隆 | 三水縣志 | 嘉慶 |
| 裕州志 | 乾隆 | 齊東縣志 | 民國 | 遷江縣志 | 民國 | 禹縣志 | 民國 |
| 南陵縣志 | 民國 | 廣信府志 | 嘉靖 | 鹽城縣志 | 光緒 | 高陵縣志 | 嘉靖 |
| 容縣志 | 光緒 | 豪州志 | 光緒 | 吳江縣續志 | 光緒 | 江都縣志(3) | 萬曆 |
| 重修滑縣志 | 民國 | 太湖縣志 | 同治 | 建平縣志 | 民國 | 興化縣新志 | 萬曆 |
| 寧夏府志 | 乾隆 | 高陽縣志 | 民國 | 滁州志 | 光緒 | 贛榆縣志 | 光緒 |
| 朔方道志(2) | 民國 | 定縣志 | 民國 | 永新縣志 | 同治 | 陵縣志 | 光緒 |
| 遷安縣志 | 民國 | 恩平縣志 | 民國 | 瀏陽縣志 | 同治 | 高陵縣續志 | 光緒 |
| 信宜縣志 | 光緒 | 大名縣志 | 民國 | 平江縣志 | 同治 | 廬陵縣志 | 民國 |
| 重修電白縣志 | 光緒 | 荊州府志(4) | 光緒 | 吉水縣志 | 光緒 | 續華州志 | 康熙 |
| 元和縣志 | 乾隆 | 黃州府志 | 光緒 | 義寧州志(2) | 同治 | 再續華州志 | 乾隆 |
| 重修毗陵志 | 成化 | 曹州府志 | 乾隆 | 饒州府志(2) | 同治 | 江寧新志 | 乾隆 |
| 嘉定縣志 | 萬曆 | 福山縣志 | 乾隆 | 安順府志(4) | 咸豐 | | |
| 吉林通志(3) | 光緒 | 嵊縣志 | 同治 | 保定府志(3) | 光緒 | | |

**2. The paper mentioned that 562 of the new records have correspondence in CBDB. It might be interesting to give some geographical analysis of the findings.**

We have changed our program since we submitted the manuscript for review. The most recent version identified 602 records that have matching records in CBDB. The following table shows the *Difangzhi* sources from which we could find these records. Since some records appeared in more than one time, the total of occurrences is more than 602.

| Source | Times | Source | Times | Source | Times | Source | Times |
|---|---|---|---|---|---|---|---|
| 順德縣志 | 4 | 豐縣志 | 2 | 祁門縣志 | 6 | 太湖縣志 | 3 |
| 陽江志 | 14 | 華亭縣志 | 8 | 慈利縣志 | 0 | 續修舒城縣志 | 1 |
| 臨晉縣志 | 0 | 青浦縣志 | 12 | 長沙府志 | 9 | 廣西通志輯要 | 103 |
| 漢州志 | 8 | 樂昌縣志 | 1 | 襄城縣志 | 0 | 三水縣志 | 0 |
| 裕州志 | 0 | 齊東縣志 | 2 | 遷江縣志 | 0 | 禹縣志 | 11 |
| 南陵縣志 | 3 | 廣信府志 | 13 | 鹽城縣志 | 4 | 高陵縣志 | 2 |
| 容縣志 | 10 | 豪州志 | 25 | 吳江縣續志 | 1 | 江都縣志 | 4 |
| 重修滑縣志 | 23 | 太湖縣志 | 4 | 建平縣志 | 1 | 興化縣新志 | 0 |
| 寧夏府志 | 27 | 高陽縣志 | 6 | 滁州志 | 15 | 贛榆縣志 | 1 |
| 朔方道志 | 31 | 定縣志 | 13 | 永新縣志 | 8 | 陵縣志 | 2 |
| 遷安縣志 | 2 | 恩平縣志 | 9 | 瀏陽縣志 | 3 | 高陵縣續志 | 2 |
| 信宜縣志 | 0 | 大名縣志 | 32 | 平江縣志 | 4 | 廬陵縣志 | 13 |
| 重修電白縣志 | 6 | 荊州府志 | 45 | 吉水縣志 | 4 | 續華州志 | 0 |
| 元和縣志 | 0 | 黃州府志 | 37 | 義寧州志 | 8 | 再續華州志 | 0 |
| 重修毗陵志 | 23 | 曹州府志 | 11 | 饒州府志 | 75 | 江寧新志 | 7 |
| 嘉定縣志 | 0 | 福山縣志 | 1 | 安順府志 | 3 | | |
| 吉林通志 | 1 | 嵊縣志 | 8 | 保定府志 | 27 | | |

**3. The extracted records seem to focus on officials. However, gazetteers also discuss people who were important to local history, such as those who passed exams, virtuous women, and local dignitaries. These are the people who may have left more significant local impact than the officials. It might be interesting to include them as well.**

We will strengthen our programs to expand the types and coverage of biographical information gradually.

## Reviewer 2

**Can your method identify persons definitely? Are there exceptions? Are there sentences that your filter patterns could not be applied?**

No, we are not completely sure that the identified records definitely correspond to particular persons in CBDB. As we discussed in the paper, we employed contextual information to disambiguate. Hence, the chances of mistakenly matching records for two different persons are low. It demands an exceptionally unfavorable condition for the dynasty, name, and zi of a person recorded in *Difangzhi* to incorrectly match a record in CBDB. The chance of mis-matching is even reduced now since we are able to identify a person's hao (號) in *Difangzhi*. However, we have not manually examined all of the identified and matched records yet. We are still expanding the mechanism for automatic tagging of *Difangzhi* books, and will verify the results of automatic tagging at a later time.

Yes, there are sentences that the current patterns could not be applied. The word orders of Mandarin Chinese are quite flexible, so it is possible to encounter relatively uncommon sentence patterns in *Difangzhi* that our patterns cannot handle. In such cases, we may choose to add a pattern, either manually or algorithmically. However, adding a filter pattern for relatively uncommon word orders may help the recall rate and hurt the precision rate at the same time.

## Reviewer 3

**1. I could only recommend the authors to describe their methods in full detail in a fashion that scholars unfamiliar with the technology can understand it as well.**

We are sorry that we could not describe our methods completely due to the 1500-word constraint for DH 2015 papers. We plan to do so in an extended paper in the near future.

**2. Please give some clear examples of matching historical data and new found historical data. So not just statistics, but a few readable examples that will convince historians how worthwhile your methods are.**

Two examples are provided below.

The first instance shows the potential benefits when we can identify biographical information that matches some CBDB records.

Figure A1 shows a *Difangzhi* paragraph for a person whose name (李滋然), zi (命三), and dynasty (清). Our programs extracted a record "清,李滋然,命三" from a text version of this image (the third line from the right end of this page). Although the text does not specify the dynasty "清" explicitly, the context provides unambiguous hints, i.e.,, "光緒". As a result, we could match this record with the CBDB person ID 77918. Furthermore, the paragraph may provide additional information about "李

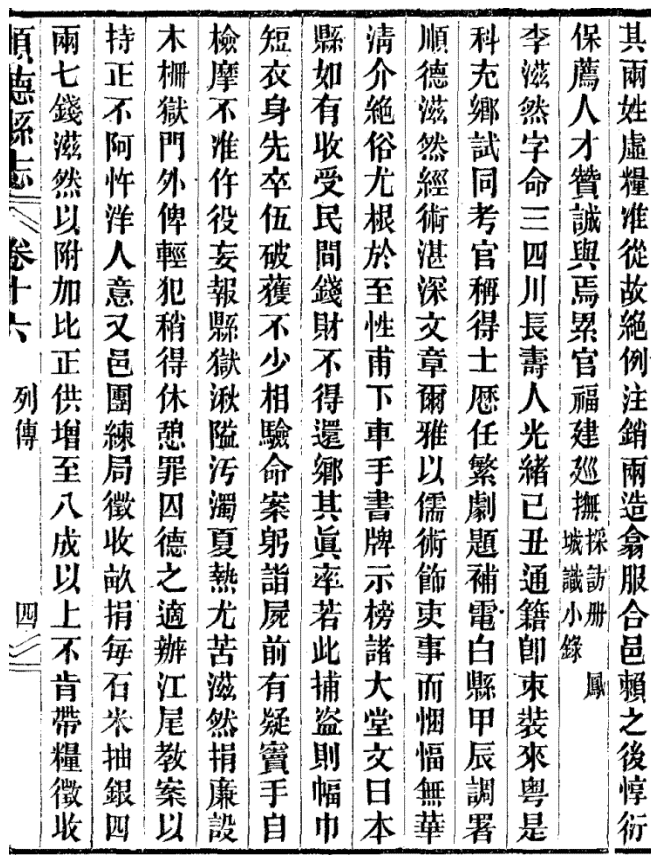

Figure A1. A *Difangzhi* paragraph for 李滋然

滋然" that we can use to enrich the CBDB database.

We use Figure A2 to illustrate the potential contribution of an incomplete match between a record that we identified from *Difangzhi* and the CBDB records.

From a text version of the image in Figure A2 (leftmost paragraph on the page), our programs identified a name (薛平) and the accompanying zi (坦途). Given the mentioning of "節度使", we judged that the person belonged to the Tang dynasty. Hence, we have a record "唐,薛平,坦途".

When we attempted to match this record with the CBDB records, we could only find person IDs 180112 and 180316. Both records are for a "薛平" in the Tang dynasty, but neither contains information about zi and hao for "薛平".

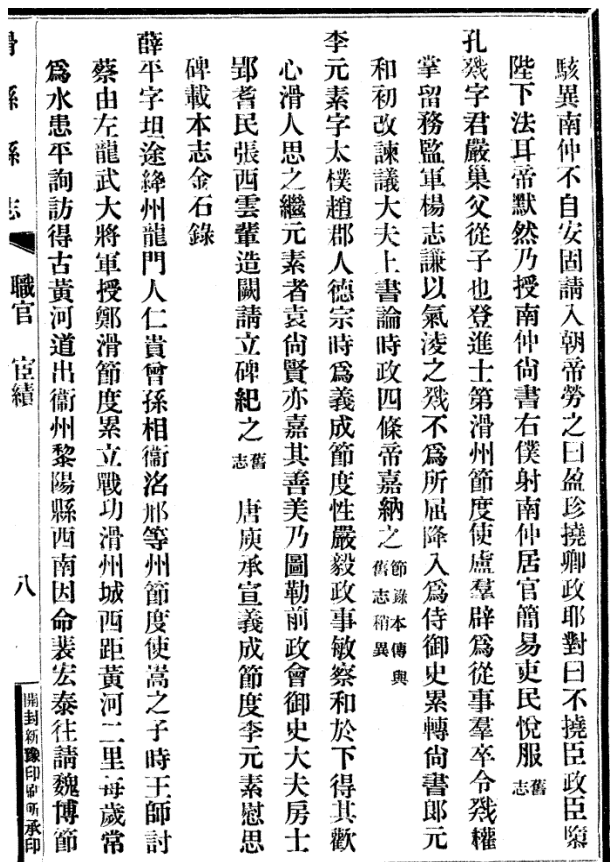

Figure A2. A *Difangzhi* paragraph for 薛平

Hence, the information, zi (坦途) in this case, that we extracted from *Difangzhi* might be useful for adding unknown information to the CBDB database, if the match between these records can be verified and confirmed by experts. In addition, as we explained for the first case, the leftmost paragraph might provide extra biographical information about "薛平".

## Reviewer 4

**Given that the authors admit that their results need to be interpreted and confirmed by a domain expert to verify their accuracy, I was somewhat surprised at the lack of technical details found in this proposal. While I applaud them for writing to a non-specialized audience, at least some detail as to the nature of the algorithms and the systems that they are using would have been appreciated and the absence of this information (which I'm sure the authors could provide if this is accepted) gives me pause. ... I would have preferred to see a more developed and complete proposal submitted as a long paper.**

There is no denial that not all of the technical details were provided in the proposal. We will provide all details when we present our work at the conference, and look forward to publishing the details in an extended report.

We attempted to include a clear outline of our technical methods, using more than 500 words in a 1500-word abstract. To make a complete extended abstract, we had no choice but using about 1000 words for background, problem definition, evaluation results, and discussions.